\documentclass{article}

\usepackage{microtype}
\usepackage{graphicx}
\usepackage{subfigure}
\usepackage{booktabs} 
\usepackage{textcomp}

\usepackage{hyperref}

\usepackage[accepted]{icml2018}

\icmltitlerunning{Achieving Fairness through Adversarial Learning: an Application to Recidivism Prediction}

\begin{document}

\twocolumn[
\icmltitle{Achieving Fairness through Adversarial Learning: an Application to Recidivism Prediction}




\author{
  \Large{Christina Wadsworth}\\
  \normalsize{Stanford University}\\
  \normalsize{Stanford, CA}\\
  \normalsize{cwads@cs.stanford.edu}
  \and
  \Large{Francesca Vera}\\
  \normalsize{Stanford University}\\
  \normalsize{Stanford, CA}\\
  \normalsize{fvera@cs.stanford.edu}
  \and
  \Large{Chris Piech}\\
  \normalsize{Stanford University}\\
  \normalsize{Stanford, CA}\\
  \normalsize{piech@cs.stanford.edu}
}
\title{}
\date{\vspace{-5ex}}
\vspace{-8ex}
\maketitle
\vspace{-2ex}




\icmlkeywords{Machine Learning, ICML, Adversarial Learning, Fairness, Debias, Biased, COMPAS, recidivism}

\vskip 0.3in
]




\let\thefootnote\relax\footnotetext{FAT/ML Workshop, July 2018, Stockholm, Sweden.}

\begin{abstract}
\vspace{-1mm}
Recidivism prediction scores are used across the USA to determine sentencing and supervision for hundreds of thousands of inmates. One such generator of recidivism prediction scores is Northpointe\textquotesingle s Correctional Offender Management Profiling for Alternative Sanctions (COMPAS) score, used in states like California and Florida, which past research has shown to be biased against black inmates according to certain measures of fairness. To counteract this racial bias, we present an adversarially-trained neural network that predicts recidivism and is trained to remove racial bias. When comparing the results of our model to COMPAS, we gain predictive accuracy and get closer to achieving two out of three measures of fairness: parity and equality of odds. Our model can be generalized to any prediction and demographic. This piece of research contributes an example of scientific replication and simplification in a high-stakes real-world application like recidivism prediction.

\end{abstract}

\vspace{-5mm}

\section{Introduction}
\label{introduction}
\vspace{-2mm}

Machine learning models and other data-based algorithms are being used increasingly in decision-making processes that affect individual lives. No longer is accuracy the only concern when developing models -- fairness must be taken into account as well. Criminal risk assessment \cite{Angwin1}, salary prediction \cite{BBC}, and loan approvals \cite{Swarns} are examples of cases where existing societal biases against a certain gender or race can be perpetuated by machine learning or other algorithms and existing discrimination.

Much of recent discussion surrounding fairness has revolved around Correctional Offender Management Profiling for Alternative Sanctions (COMPAS), a risk assessment and recidivism prediction score developed by Northpointe and used widely across the United States to affect sentencing and supervision on an individual level. As it stands, COMPAS is essentially a ``black box" since Northpointe has not publicly released their algorithm, only hinting at important factors and the use of part of a 137-question survey that includes sections such as: Criminal History, Social Environment, Criminal Personality, and Criminal Attitudes \cite{Angwin1}. Angwin et al. \yrcite{Angwin1} published a report on ProPublica that claimed COMPAS is biased against black inmates based on a study conducted using data from Florida inmates. COMPAS performs similarly in terms of accuracy for white and black inmates, but the errors COMPAS makes indicate discrimination against black inmates, according to Angwin et al. For example, a black inmate who does not re-offend is more likely to be classified as ``high risk" than a white inmate who does re-offend. Black inmates in general are almost three times more likely to be classified as ``high risk" than white inmates. Northpointe responded to the ProPublica analysis by arguing that COMPAS satisfies the fairness metric of calibration \cite{Equivant}.

Bias against black inmates can be learned in neural networks that predict recidivism as well. Even if race is not an input feature, other features are correlated with race. For example, an inmate's number of priors and previous time in jail are correlated with race because black inmates are more likely to be jailed for the same crimes as white inmates and are arrested at a higher rate for less serious crimes \cite{Williams, Fenton}.


\textbf{Contribution} By adding an adversary to a network that predicts recidivism, we counteract racial biases found in criminal history datasets. Our adversary penalizes our recidivism prediction network if race is predictable from the recidivism prediction. Our model is generalizable to almost any prediction and any demographic.

\section{Related Work}
\vspace{-2mm}
\textit{Fairness Definitions} The three types of fairness often used in fairness research, which are also used in this paper, are: demographic parity, equality of odds, and calibration. Hardt et al. \yrcite{HardtPS16} and Kleinberg et al. \yrcite{Kleinberg} provide definitions for these types of fairness and discuss their trade-offs. In the context of predicting recidivism, they are defined:
\newline \textbf{Parity} A score $S = S(x)$ satisfies parity if the proportion of individuals classified as high-risk is the same for each demographic. 
\newline \textbf{Equality of Odds} A score $S = S(x)$ satisfies equality of odds if the proportion of individuals classified as high-risk is the same for each demographic, when true future recidivism is held constant. White and black inmates that do recidivate should have the same proportion of high risk classification.
\newline \textbf{Calibration} A score S = S(x) is calibrated if it reflects the same likelihood of recidivism irrespective of the individual\textquotesingle s demographic. In this application, black inmates who are classified as high risk should have the same probability of true recidivism as white inmates classified as high risk.

\textit{Adversarial Fairness} Past research has explored the use of adversarial networks to achieve fairness. Beutel et al. \yrcite{Beutel} used an adversary on a shared hidden layer to satisfy parity for salary prediction, and Ganin et al. \yrcite{Ganin} used a domain classifier and reverse gradient on a hidden layer to remove domain correlation. Recently, Zhang et al. \yrcite{Zhang} used a predictor and adversary with an additional projection term to satisfy parity or equality of odds with word embeddings and predicted salary. Zhang et al. are the only existing work that uses the output layer of their predictor as input to the adversary, which was a departure from the conventions we noted earlier. No published adversarial research thus far has looked at recidivism. We replicate and simplify Zhang et al.\textquotesingle s model to explicitly demonstrate how adversarial models that achieve fairness can have real-world applications.

\textit{Fairness and COMPAS} There is much debate among researchers who have studied COMPAS as to which definitions of fairness should be satisfied for recidivism predictions. Corbett-Davies et al. \yrcite{Corbett-DaviesP17} argue that because black inmates are arrested at a higher rate than white inmates, there should be some disparity of recidivism prediction for white and black inmates. Other researchers investigate COMPAS in relation to priors since black inmates have more priors \cite{Chouldechova}. We challenge these notions: if black people are more likely to become incarcerated in the US than white people when controlling for criminal behavior \cite{Bridges}, black inmates should not be punished for our biases with harsher recidivism predictions.

\textit{Methods to Improve Fairness on COMPAS} Past research has attempted to improve racial fairness within COMPAS according to the metrics of false positive and false negative differences across black and white inmates. Hardt et al. \yrcite{HardtPS16} used post-processing on standard logistic regressor that picks different thresholds for different groups and at times adds randomization. Zafar et al. \yrcite{ZafarR17} first tried to enforce equal False Positive and False Negative rates for groups by introducing penalties for misclassified data points during training. They then optimized by using the covariance between sensitive attributes and distance between feature vectors of misclassified samples and classifier boundary. Bechavod et al. \yrcite{BechavodL17} used a logistic regressor while simultaneously penalizing unfairness by penalizing the difference in the average distance from the model\textquotesingle s decision boundary across values of the protected attribute. They use an Absolute Value Difference (AVD) model and a Squared Difference (SD) model. We compare our model to these existing methods that aim to improve fairness for COMPAS. \\
Other research has been done to debias COMPAS, focusing on debiasing input features. Lum et. al. propose univariate transformations of input features to achieve independence from the protected demographic, then use random forest to predict unbiased outputs \yrcite{JohndrowLum2016}. Johndrow et. al. apply this model to recidivism prediction \yrcite{JohndrowLum2017}. Ludrow et. al. transform the input dataset so that predictability of the protected demographic is impossible before classification. We contribute to this work by creating a model that optimizes for an unbiased output, as opposed to optimizing for unbiased input features. Additionally, we present separate models to satisfy multiple definitions of fairness.
\vspace{-2mm}
\section{Adversarial Model}

\subsection{Model Structure} 
\vspace{-3mm}
We start with a multi-layer neural network $N$ that outputs a probability of recidivism, denoted $\hat{Y}$. $\hat{Y}$ is our primary prediction objective, and $N$ is our baseline. We now want our output $\hat{Y}$ to satisfy demographic parity or equality of odds for demographic $D$. Even if $D$ is not an input feature to our neural network, it may be correlated with other features, from which the network can learn a bias. 
\newline\textbf{Demographic Parity Model}: We input the $logit$ from $N$ (the unnormalized predicted recidivism probability, i.e. just before the sigmoid) to an adversarial neural network $A$ that learns to classify demographic $D$. If $\hat{Y}$ is biased for demographic $D$, $A$ should learn to have a high accuracy because the $logit$ will be highly predictive of $D$. Our goal is for neural network $N$ to predict $\hat{Y}$ accurately and for $A$ to predict $D$ poorly. \newline\textbf{Equality of Odds Model}: We can input the $logit$ as well as the true recidivism value $Y$. Here the goal is for there to be no difference in $\hat{Y}$ across demographic, given true recidivism value $Y$, which will be satisfied when $A$ cannot predict $D$ with high accuracy given $Y$ and the $logit$.


\begin{figure}[ht]
\vskip 0.2in
\begin{center}
\centerline{\includegraphics[width=1\columnwidth]{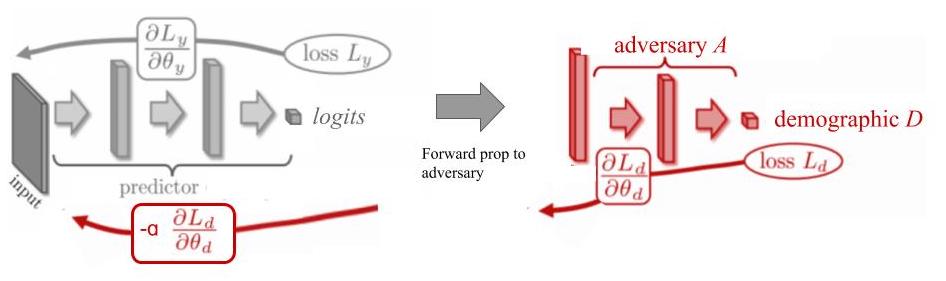}}
\caption{Diagram of our adversarial model structure.}
\label{model_structure}
\end{center}
\vskip -0.2in
\vspace{-5mm}
\end{figure}

\vspace{-3mm}
\subsection{Model Training} 
\vspace{-3mm}
Our goal is for neural network $N$ to predict $\hat{Y}$ accurately and for $A$ to predict $D$ poorly. If we achieve this, our model will be outputting an unbiased, accurate $\hat{Y}$. More specifically, we use binary cross-entropy losses for $N$ and $A$, which we refer to as $L_y$ and $L_d$, respectively. To train $A$, we back-propagate $L_y$ through $A$. However, we need to train $N$ to be good at predicting $\hat{Y}$ and bad at predicting a $logit$ that is highly correlated with $D$. If we subtract $L_d$ from $L_y$, $N$ will be encouraged to maximize $L_d$, which will produce a $logit$ that cannot be used to predict race and $\hat{Y}$ values that are closer to achieving parity. We train our model $N$ with the following loss function:
\\
\\
\centerline{$L = L_y - \alpha*L_d$}
\vspace{-6mm}
\section{Experiments and Results}
\vspace{-2mm}
\textit{Data} We apply our adversarial model to recidivism prediction. To do so, we used public criminal records data from Broward County, Florida that was compiled and published by ProPublica. Much of recidivism research in the past two years has been conducted on this dataset. The dataset also includes COMPAS scores for Broward County inmates, so we are able to compare our results to the performance of COMPAS. Our training set is size 8230 and our test set size 2213. We only use data from white and black inmates, of which 41\% represents white inmates and 59\% black inmates. Beutel et al. \yrcite{Beutel} showed that a more obviously skewed distribution on demographic $D$ can affect how helpful the adversary is. We found that despite our slight skew, the adversary was just as effective.

\textit{Model} Predictor $N$ has 2 256-unit ReLU hidden layers. Adversary $A$ has a single 100-unit ReLU hidden layer. We used a learning rate of $e^{-4}$, binary cross entropy loss, a sigmoid output layer, an Adam optimizer, and an alpha value of 1. To settle on these hyper parameters, we tuned a number of hidden layers and hidden layer size for both the predictor and adversary, alpha, and learning rate. Figure 2 shows tuning for number and size of hidden layers for $N$. When tuning alpha, we wanted to maximize $L_d$, then minimize $L_y$. This allows for fairness to be satisfied first, then for our predictor $N$ to be as accurate as possible.

\textit{Metrics} To evaluate accuracy, we use area under ROC curve. We also define metrics that can be used to compare demographic parity and equality of odds:
\\ \\
\vspace{-3mm}
\centerline{\textit{High Risk Gap: $|{High Risk_{white} - High Risk_{black}}|$}} \\ \\
\vspace{-3mm}
\centerline{\textit{False Positive Gap: $|{FP_{white} - FP_{black}}|$}} \\ \\
\vspace{-3mm}
\centerline{\textit{False Negative Gap: $|{FN_{white} - FN_{black}}|$}} \\

\vspace{-6mm}
If High Risk Gap is zero, demographic parity is satisfied. False Positive Gap and False Negative Gap are used to assess equality of odds. If both are zero, equality of odds is satisfied. Also included are conditional probability graphs to compare evaluations on the fairness metric calibration.

\begin{figure}
\begin{center}
\includegraphics[width=.43\columnwidth,angle=270,origin=c]{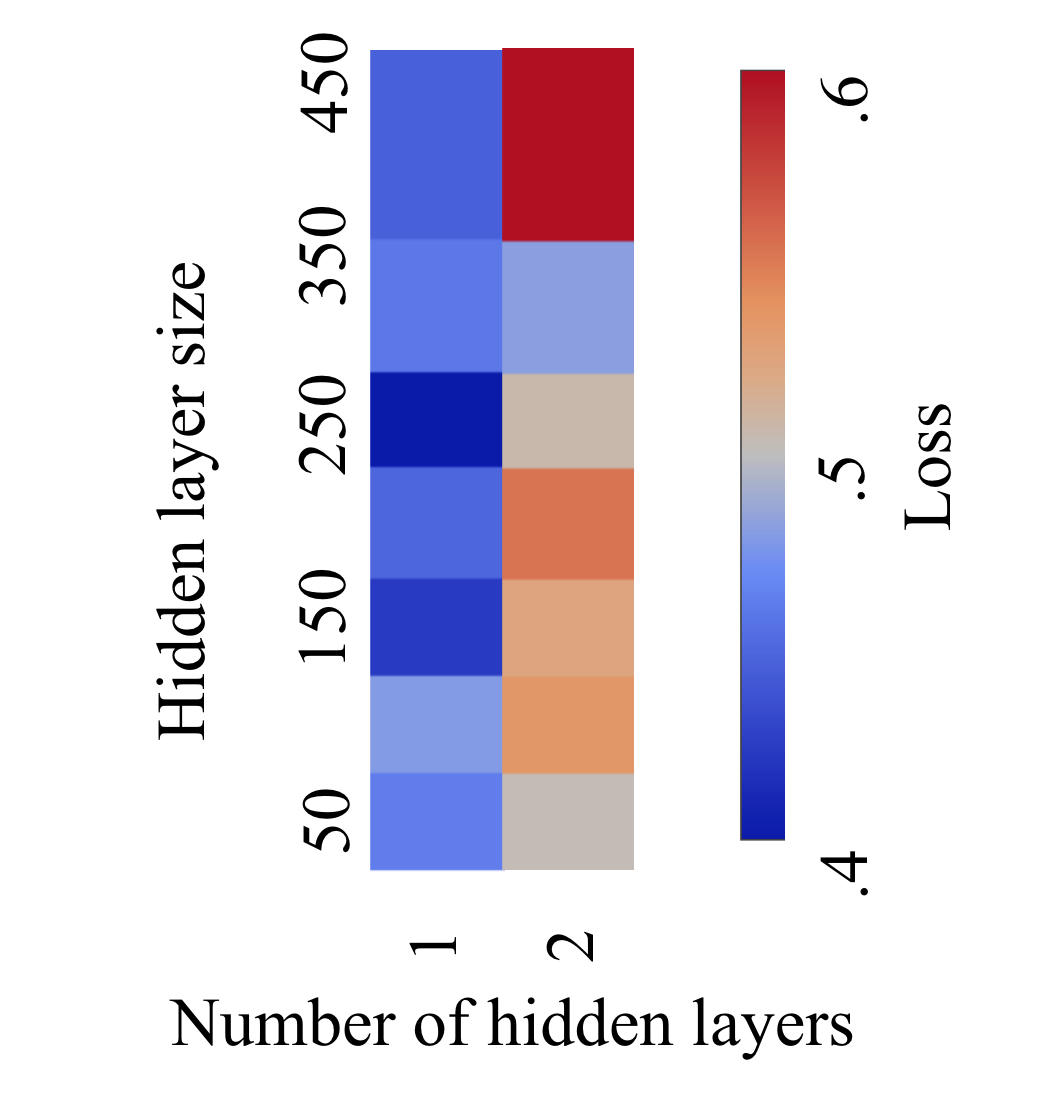}
\vspace{-5mm}
\caption{Model structure tuning.}
\label{num_layers}
\end{center}
\vspace{-5mm}
\end{figure}

\textit{Experiments} Outlined are the models used to conduct experiments on our COMPAS dataset for recidivism prediction:
\newline\textbf{Recidivism Prediction}:
We trained a regular recidivism predictor without any type of adversary to compare our adversarial models to a baseline and to confirm that bias is perpetuated in machine learning models.
\newline\textbf{Adversarial Models}:
We then trained two variants of the adversarial recidivism predictor. One adversary accepted the $logit$ as input and the other accepted the $logit$ and true recidivism value $Y$. We tried an additional variation of the model that accepted a hidden layer as input instead of the $logit$. This model was less stable and required more hyper parameter tuning to see a decent result. As such, we continued most of our research with our models that accept the $logit$ because the adversary was just as powerful and less finicky. In our results section, we present the model that accepted just the $logit$ as input as our adversarial model.

\begin{table}[h!]
\vskip 0.15in
\begin{center}
\begin{small}
\begin{sc}
\scalebox{0.75}{
\begin{tabular}{lcccr}
\toprule
Model& High Risk Gap & FN Gap & FP Gap\\
\midrule
COMPAS Scores (Our Test Set)   & 0.18&0.22&0.17\\
Our Recidivism Model   & 0.21&0.27&0.15\\
Our Chosen Adversarial Model   & \textbf{0.02}&\textbf{0.02}&\textbf{0.01}\\
\bottomrule
\end{tabular}}
\end{sc}
\end{small}
\end{center}
\vskip -0.1in
\caption{Comparison of High Risk Gap, False Negative Gap (FN Gap), and False Positive Gap (FP Gap), across models.}
\vspace{-3mm}
\end{table}

\begin{table} [h!]
\vskip 0.15in
\begin{center}
\begin{small}
\begin{sc}
\begin{tabular}{lcccr}
\toprule
Model& AUC\\
\midrule
COMPAS Scores (Our Test Set) & 0.66\\
Our Recidivism Model & 0.72\\
Our Chosen Adversarial Model & 0.70\\
\bottomrule
\end{tabular}
\end{sc}
\end{small}
\end{center}
\vskip -0.1in
\caption{Comparison of AUC for ROC curves across models.}
\vspace{-1mm}
\end{table}

\textit{Results} Our regular recidivism predictor is similarly biased against black inmates with respect to parity and equality of odds, which suggests that biases can be learned and perpetuated by machine learning models. However, the results in Table 1 show that our chosen adversarial model is much closer than COMPAS to satisfying parity and equality of odds. Further, from Table 2, we see that our adversarial model has improved accuracy compared to COMPAS scores. On the other hand, Figure 3b demonstrates that COMPAS satisfies calibration, whereas some bias is evident in our model (Figure 3a), especially at the threshold cutoff 0.5, which is the cutoff for recidivism prediction -- although our bias is slight.


When comparing feature importance for our recidivism prediction baseline model and our adversarial model, we notice that the adversarial model relies more heavily on 6 of the top 10 most important features (Figure 4). This indicates that the adversarial model depends on more holistic information while the regular recidivism model is mostly dependent on charge degree, age, and priors. Recall that black inmates have more priors because black inmates are jailed for crimes more often than white inmates are \cite{Fenton}. Additionally, charge degree distributions are different across races. It is possible that relying on these three features contributes to the racial bias found in the baseline model.


We further assess our model in the context of other work done to debias COMPAS. Most existing work compares false positive and false negative values. As shown in Table 3, our adversarial model is just as good at debiasing with regards to race as state of the art work done on COMPAS is.

Overall, our model outperforms COMPAS scores in terms of accuracy and comes close to satisfying parity and equality of odds. We also achieve false positive and false negative differences that are on par with the best of fairness research done on COMPAS. However, we note that our model is slightly more biased than COMPAS with regards to calibration.

\begin{table} [t]
\vskip 0.15in
\begin{center}
\begin{small}
\begin{sc}
\scalebox{0.7}{
\begin{tabular}{lcccr}
\toprule
Model& Accuracy& FP Gap & FN Gap\\
\midrule
COMPAS Scores (Our Test Set)   &0.68& 0.17&0.22\\
Our Recidivism Model   &0.70&0.15&0.27\\
Our Chosen Adversarial Model   &0.70& \textbf{0.01}&\textbf{0.02}\\
Bechavod et al. AVD Penalizers \yrcite{BechavodL17}  &0.65& \textbf{0.02}&\textbf{0.04}\\
Bechavod et al. SD Penalizers \yrcite{BechavodL17}  &0.66& \textbf{0.02}&\textbf{0.03}\\
Bechavod et al. Vanilla Regularized \yrcite{BechavodL17}  &0.67& 0.20&0.30\\
Zafar et al. \yrcite{ZafarR17}  &0.66& \textbf{0.03}&0.11\\
Zafar et al. Baseline \yrcite{ZafarR17} &0.66& \textbf{0.01}&0.09\\
Hardt et al. \yrcite{HardtPS16} &0.65& \textbf{0.01}&\textbf{0.01}\\
\bottomrule
\end{tabular}}
\end{sc}
\end{small}
\end{center}
\vskip -0.1in
\caption{Comparison of accuracy, False Positive Gap (FP Gap), and False Negative Gap (FN Gap) across models.}
\vspace{-5mm}
\end{table}

\begin{figure}[t]
\centering
\subfigure[Our Chosen Adversarial Model]{\includegraphics[width=.47\linewidth]{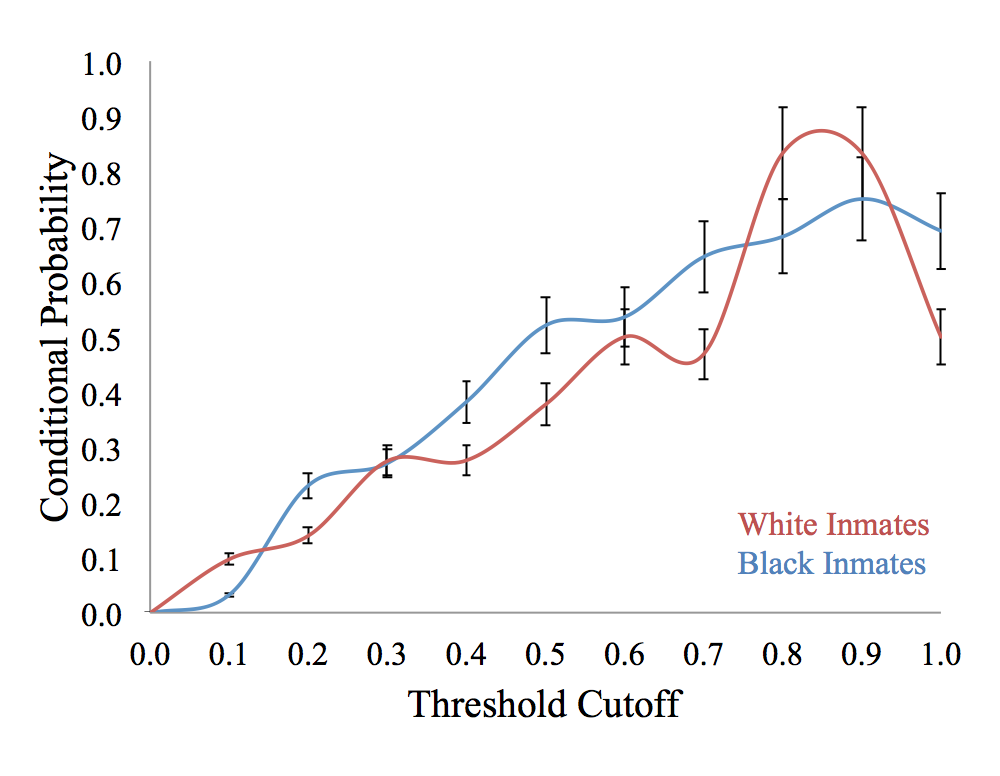}}
\subfigure[COMPAS]
{\includegraphics[width=.47\linewidth]{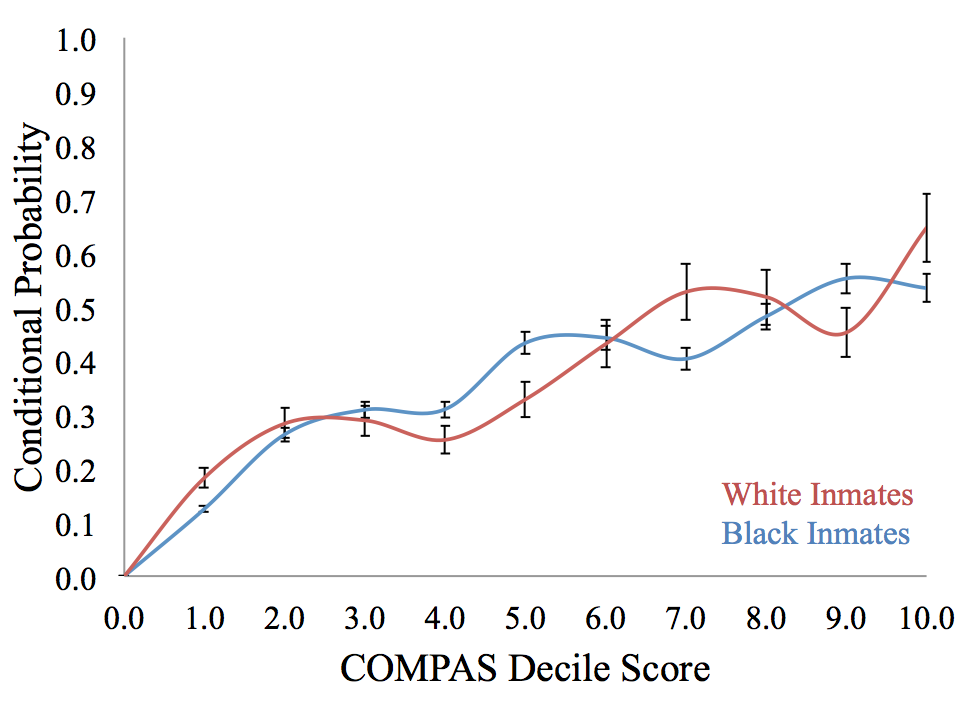}}
\caption{Conditional probability across models.}
\end{figure}


\vspace{-2mm}
\section{Case Study}
\vspace{-2mm}
The quantitative COMPAS scores impact real individuals. In a case study on two Broward County inmates, we compare COMPAS predictions to the results of our adversarial model and investigate the stories of Joe and Bob.

Despite having only 1 prior and 2 charges before COMPAS, Joe, a 55 year old black inmate, received a COMPAS score of 8 out of 10, labeling him as ``high risk" and likely to recidivate. Our adversarial model disagreed by making a prediction of 0.05 after the sigmoid layer. Presently, Joe has no charges since the COMPAS score was assigned and has not recidivated. 

In contrast, despite having 13 priors and 24 charges before COMPAS, Bob, a 27 year old white inmate, received a COMPAS score of 5 out of 10 -- a ``medium risk" score that disagrees with our prediction of 0.84 -- a score indicating Bob will recidivate. Since receiving his COMPAS score, Bob has been charged 6 times, including for disorderly intoxication and aggravated battery (an act of violent recidivism). Had our model been used, perhaps his fate would have been different due to a higher level of supervision. It is important to take into account how COMPAS and other similar models are being used to affect real lives on an individual level. A general racial bias towards black people means that individual black people are being punished.

\begin{figure}
\begin{center}
\includegraphics[width=.73\columnwidth]{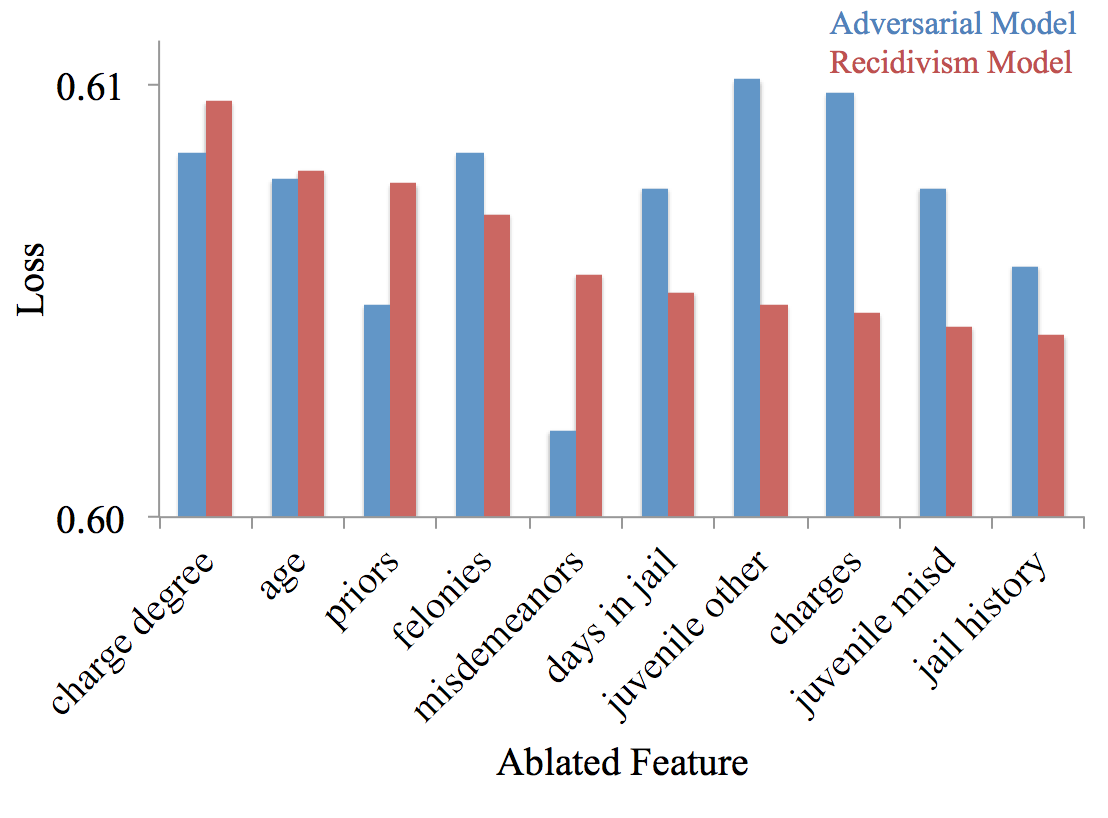}
\vspace{-3mm}
\caption{Feature importance.}
\end{center}
\vspace{-7mm}
\end{figure}

\vspace{-2mm}
\section{Conclusion}
\vspace{-2mm}
We have shown that it is possible to reduce the bias of a machine learning model that is trained on demographically biased data. We also demonstrate a general method for training unbiased models that can enforce constraints for multiple definitions of fairness. Our adversarial models are less biased than the original COMPAS scores and our recidivism prediction baseline, but the models still outperform COMPAS in terms of accuracy, providing the first piece of research on recidivism and COMPAS that achieves this with adversarial learning.


\section*{Acknowledgments}

Thanks to Ramtin Keramati for his helpful advice and Jerry Cain for his continued support of this research.

\bibliography{final}
\bibliographystyle{icml2018}

\end{document}